\documentclass[12pt]{elsarticle}
\makeatletter
\def\ps@pprintTitle{%
 \let\@oddhead\@empty
 \let\@evenhead\@empty
 \def\@oddfoot{\leftline{\footnotesize{December 2014}}}%
 \let\@evenfoot\@oddfoot}
\makeatother

\usepackage{pifont}
\newcommand{\cmark}{\ding{51}}%
\usepackage{amssymb}

\usepackage{paralist}
\usepackage{enumerate}
\usepackage{multirow}
\usepackage{arydshln} 

\usepackage{ulem} 

\usepackage{hyperref}
\hypersetup{
  colorlinks=true,
  citecolor=darkgreen,
  linkcolor=darkred,
  urlcolor=blue}
\journal{}

\begin{document}

\begin{frontmatter}



\title{Temporal ordering of clinical events}


\author{Azad Dehghan}
\address{a.dehghan@manchester.ac.uk}
\address{The University of Manchester, School of Computer Science, Manchester, UK}

\begin{abstract}
This report describes a minimalistic set of methods engineered to anchor clinical events onto a temporal space. Specifically, we describe methods to extract clinical events (e.g., Problems, Treatments and Tests), temporal expressions (i.e., time, date, duration, and frequency), and temporal links (e.g., Before, After, Overlap) between events and temporal entities. These methods are developed and validated using high quality datasets.
\sep
\end{abstract}


\end{frontmatter}



\section{Introduction}
Temporal ordering of events from semi-/un-structured textual data (e.g., news article, clinical narrative) has important applications in many practical clinical applications such as questioning and answering (e.g., personal assistance), timeline analysis (e.g., event monitoring, pathway extraction), and text summarisation. 

Chronological ordering of events involves the tasks of named entity recognition and classification (NER) or event extraction, including temporal entity recognition and normalisation (TERN), and temporal relation (TLINK) identification and classification. 

\begin{figure}[h]
\includegraphics*[scale=0.29]{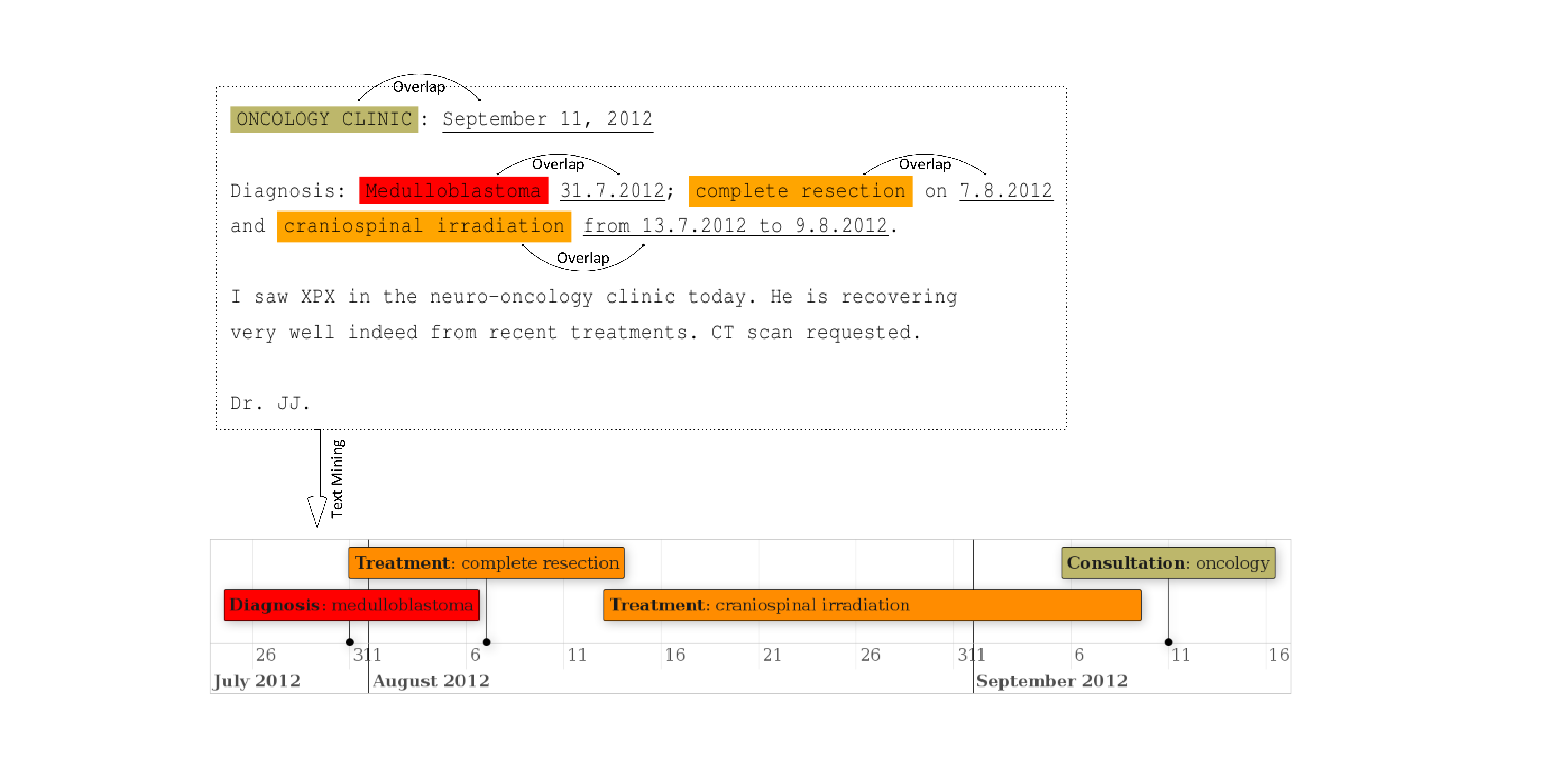} \caption{Chronologically ordered events from a sample clinical narrative} \label{figure:note_ex}
\end{figure}

Moreover, temporal ordering of events from textual clinical data include, at least, three NLP tasks: \begin{inparaenum}[(1)] \item event extraction (e.g., clinical events or concepts such as problems, treatments and tests), \item temporal entity extraction: identification (e.g., `January 4 1988', `twice daily') and normalisation, and \item temporal relations extraction (\textit{determine when a particular event occurred}).\end{inparaenum} For example, in Figure \ref{figure:note_ex} a number of events (highlighted) and TE (underlined) have been identified in a sample clinical narrative. Subsequently, the chronological ordering of events have been visualised in the given timeline.

The methods described in this report has been inspired from a number of work derived from community held evaluations in relevant NLP tasks: 

\subsubsection*{Event extraction}
Recent work in clinical event extraction has been notably pushed by recent community held evaluation in clinical named entity recognition organised as part of 2010 \cite{Uzuneretal11} and 2012 \cite{Sunetal13b} i2b2 challenges. 

\subsubsection*{Temporal entity extraction}
Likewise, temporal entity extraction has been notably pushed by a number of general domain SemEval/TempEval \cite{Verhagenetal07,Verhagenetal10,Uzzamanetal13} and notably specific domain  2012 i2b2 \cite{Sunetal13b}  challenges.

\subsubsection*{Temporal relation extraction}
The aim of temporal relation extraction is to anchor extracted events onto a temporal space. Recent work on this problem have resulted from the 2012 i2b2 \cite{Sunetal13b} and more recently SemEval-2015 task 6, Clinical TempEval \cite{Bethard2015}.\\ 

The remainder of this paper is structured as followed: Section \ref{sec:event_extraction} described the methods engineered to extract clinical events such as medical problems, treatments and tests. Section \ref{sec:temporalentity_extraction} describes the methods developed to identify and normalise (ISO-8601). Section \ref{sec:tlink_extraction} describes the temporal entity identification and classification approach. Section \ref{sec:experiments_results_discussions} presents the experiments, results and dicussions. The conclusion is given in the final Section \ref{sec:conclusion}.

This paper is largely self-contained\\ 

Note that this report is a reprint from the author's thesis [TBA] and significant improvement of intermediate results previously published \cite{Kovacevicetal13}.\\

A number of components described herein are available as open source\footnote{Clinical NERC \url{http://sourceforge.net/projects/clinical-nerc/} and Clinical TERN \url{http://sourceforge.net/projects/clinical-tern/}}\\

\section{Event extraction}
\label{sec:event_extraction}
The aim of the event extraction method is to identify broad clinical event categories such as, \textit{Problem}, \textit{Treatment} and \textit{Test} and map them to a medical knowledge base such as the UMLS Metathesaurus for fine-grained semantic characterisation of event instances\footnote{No evaluation is provided on event/concept mapping.}. These event categories will collectively be referred to as EVENTs from henceforth. We have adopted the i2b2 definitions of concept or event categories which are largely based on the UMLS semantic types, but not limited by their coverage\footnote{\url{https://www.i2b2.org/NLP/Relations/assets/Concept\%20Annotation\%20Guideline.pdf}}; 

\subsection{Methods}
The core NER is a data-driven approach (using the state-of-the-art sequence labelling algorithm CRF) to identify clinical EVENTs from healthcare narratives.

The EVENT extraction pipeline is made up of three main processing components: NLP pre-processing, the NER (see Figure \ref{figure:ClinicalNER}), and Negation.:

\begin{figure}[h]
\begin{center}
\includegraphics*[scale=0.5]{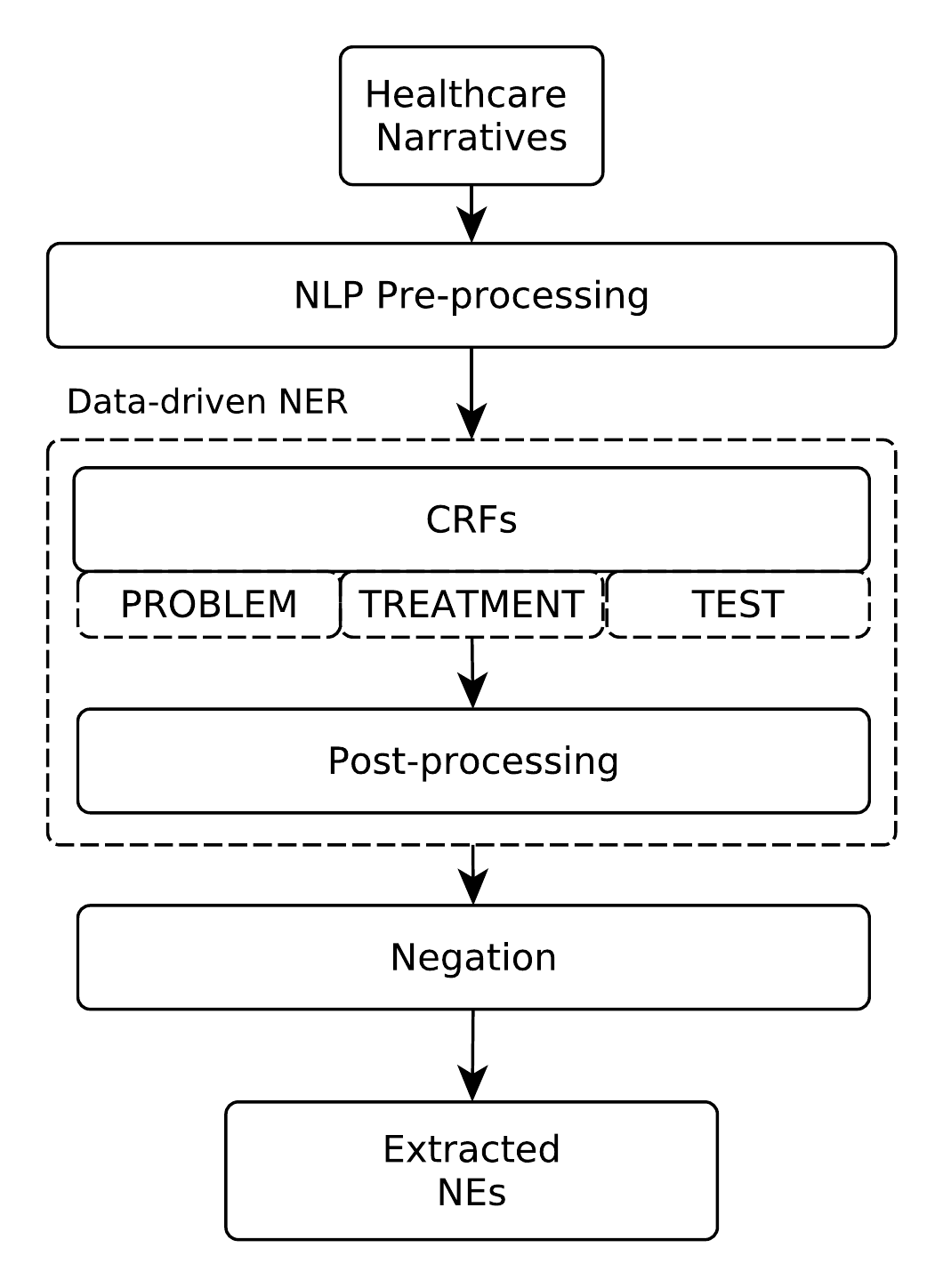} \caption{EVENT extraction architecture} \label{figure:ClinicalNER}
\end{center}
\end{figure}

The NLP pre-processing pipeline is made up of lexical and syntactic processing components, specifically: \begin{inparaenum}[(1)] \item Tokeniser, \item sentence splitter, \item word stemmer, \item POS tagger, and \item chunker / shallow parser.\end{inparaenum}

\subsubsection*{Data-driven NER}
The Data-driven NER component utilises separate CRFs trained for each EVENT category: \textit{Problem}, \textit{Treatment} and \textit{Test}. A combination of the forward and backward feature selection approaches were adopted to select a total of 20 most discriminant features from an initial set of 120 features. The same set of features were used across all categories as our analysis showed this was the best fit. The extracted features can be clustered into two sets: lexical and syntactic, with four feature groups across (see the below list).

\begin{itemize}
 \setlength{\itemsep}{1pt}
 \setlength{\parskip}{0pt}
 \setlength{\parsep}{0pt}
\item[\textbf{Lexical}] 
\item $fg_1$: the token string or alphanumeric character sequence
\item $fg_2$: the stem of each token
\item $fg_3$: the POS-tag for each token 
\item[\textbf{Syntactic}]
\item $fg_4$: the chunk tag for each token
\end{itemize}

Further, the feature space is also made up of contextual features of the neighbouring tokens with a feature window size of 5 or [-2,2] with respect to the current token. The \textit{window size} corresponds to the number of tokens to the \textit{left} and \textit{right}, including the current token, of which contextual token features are considered. Specifically, for each token $t$ and a given feature group $fg$, the feature space consists of: ($t_{fg}$), ($t_{fg}$+1), ($t_{fg}$+2), ($t_{fg}$-1), and ($t_{fg}$-2) (see Table \ref{table:NERtemplate}).

\begin{table}[h] \caption{Feature template: clinical EVENTs} \label{table:NERtemplate}
\begin{small}
\textit{CRF feature template used for all EVENT categories: Problem, Treatment and Test.}
\end{small}
\begin{center}
\begin{tabular}{llll} \hline
\textbf{$fg_1$:Token} & \textbf{$fg_2$:Stem} & \textbf{$fg_3$:POS} & \textbf{$fg_4$:Chunk}\\
U00:\%x[-2,1] & U05:\%x[-2,2] & U10:\%x[-2,3] & U15:\%x[-2,4]\\
U01:\%x[-1,1] & U06:\%x[-1,2] & U11:\%x[-1,3] & U16:\%x[-1,4]\\
U02:\%x[0,1]  & U07:\%x[0,2] & U12:\%x[0,3] & U17:\%x[0,4]\\
U03:\%x[1,1]  & U08:\%x[1,2] & U13:\%x[1,3] & U18:\%x[1,4]\\
U04:\%x[2,1]  & U09:\%x[2,2] & U14:\%x[2,3] & U19:\%x[2,4]\\ \hline
\end{tabular}
\end{center}
\end{table}

All CRFs were trained using a mix of BIO and W-BIO (W: single word, B: beginning, I: inside, O: outside) sequence label models with the following (default) CRF parameters: $C=1.00$, $ETA: 0.0001$ and L2-regularisation algorithm.

The post-processing component contains three sub-components:

\begin{enumerate}
 \setlength{\itemsep}{1pt}
 \setlength{\parskip}{0pt}
 \setlength{\parsep}{0pt}
\item \textbf{Label fixer}\\
This components corrects sequence label prediction from the NER component. These corrections are simple heuristics based on commonly observed errors in the training data set. Table \ref{table:labelfixer} list the full set of heuristics utilised.

\begin{table}[h] \caption{Label fixer heuristic} \label{table:labelfixer}
\begin{center}
\begin{tabular}{|c|l|l|} \hline
& \textbf{Raw predictions} & \textbf{Corrected predictions} \\ \hline
a & ... O O O I I I I ... & ... O O \textbf{B} I I I I ... \\ \hline
b & ... O O O B O O O ... & ... O O O B \textbf{I} O O ... \\ \hline
c & ... O O O B O I I ... & ... O O O B \textbf{I} I I... \\ \hline
d & ... O O O B I I B I I ... & ... O O O B I I \textbf{I} I I... \\ \hline
\end{tabular}
\end{center}
\end{table}

\item \textbf{Boundary adjustment} \\
This component attempts to expand the event boundary by including contextual tokens to the right and left of predictions that possess POS/chunk tags that corresponded to nouns and noun phrases and their constituents including adjectives and determiners (e.g., `a'; `this'; `her'). This sub-component is useful when the NER only tags part of an event. For example, if the NER component annotates the word `severe', `stomach', or `ache' from the actual term `severe stomach ache', this component would hypothetically capture the latter complete term boundary.

\item \textbf{False positive filter}\\
This component removes common false positives predictions observed during the validation of the NER, i.e., common model prediction errors. Examples of false positives prediction include single character predictions (e.g., `a'), pronouns (e.g., `he'; `she'), and determiners (e.g., `the').
\end{enumerate}

\subsubsection*{Negation}
To identify negated clinical EVENTs we used the ConText negation tool as described in \citep{Harkemaetal09}.

\section{Temporal entity extraction}
\label{sec:temporalentity_extraction}
The TERN task involves the recognition and normalisation of TEs. TE are defined by TIMEX3 schema are grouped into four temporal types: \textit{Date} (e.g., `August 23, 1993'), \textit{Time} (e.g., `2:23 p.m.'), \textit{Frequency} (e.g., `every morning'), and \textit{Duration} (e.g., `two weeks'). In addition, the \textit{Date and time format: ISO-8601} standard is used to normalise TEs into a standardised format.

\subsection{Methods} 
We propose a hybrid-based TER component, with a rule-based temporal normalisation component (ClinicalNorMA)\footnote{https://github.com/filannim/clinical-norma}. The motivation for adopting a hybrid approach for TER was to compare different approaches, and potentially combine the methods for the best possible performance.

\subsection*{Architecture}
\begin{figure}[h] 
\begin{center}
\includegraphics[scale=0.5]{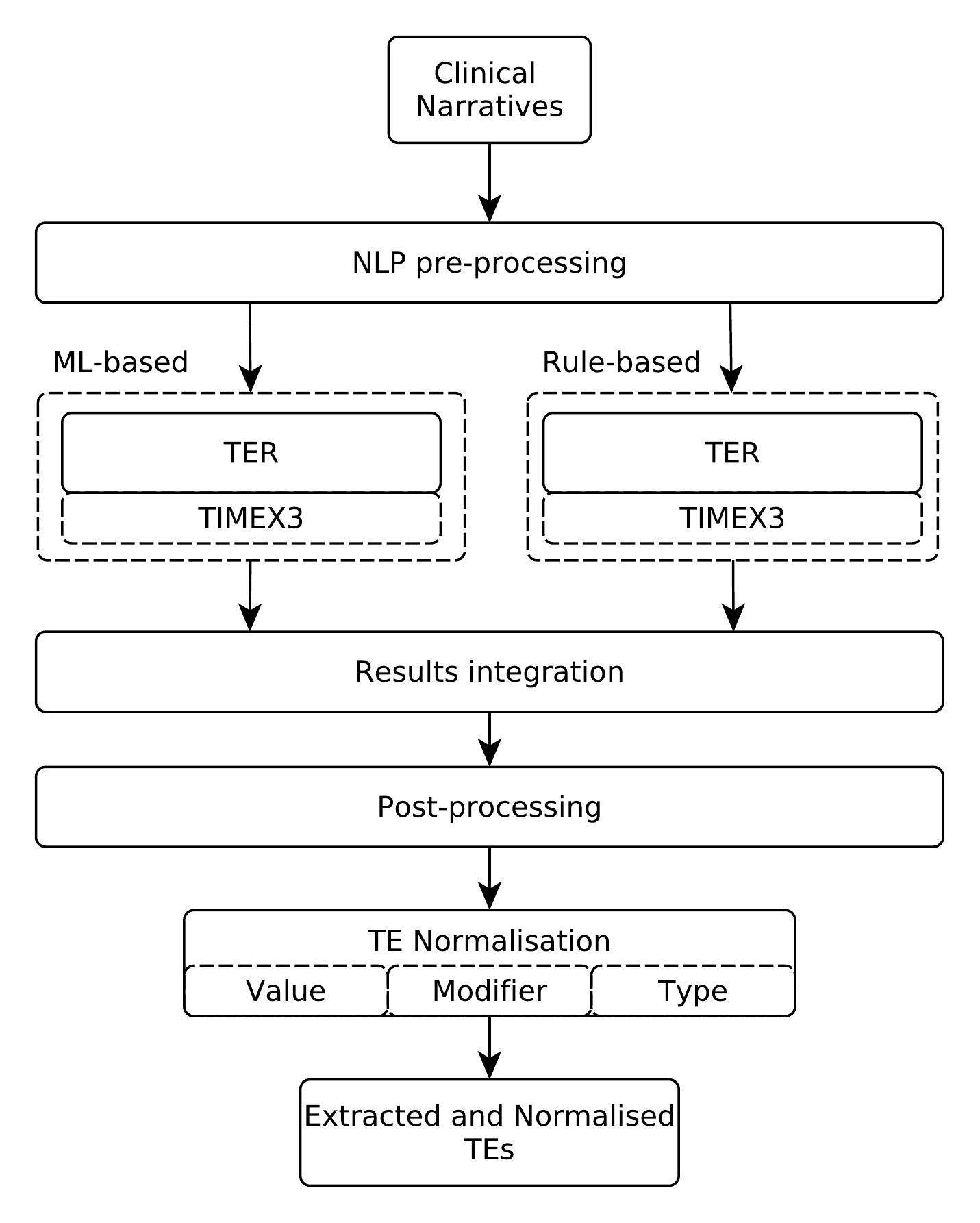} \caption{TERN architecture} \label{figure:TERNArchitecture}
\end{center}
\end{figure}

The TERN component is made up of the following components (see Figure \ref{figure:TERNArchitecture} for a overview).

A pre-processing pipeline is made up of the following NLP components: \begin{inparaenum}[(1)] \item tokeniser, \item sentence splitter, and \item semantic temporal resources.\end{inparaenum} Specifically, several bespoke temporal knowledge resources were manual compiled and applied at this stage of processing to subsequently be utilised as features for the rule- and ML-based TER components. These semantic resources cover a broad set of temporal expression sub-categories:

\begin{itemize}
 \setlength{\itemsep}{1pt}
 \setlength{\parskip}{0pt}
 \setlength{\parsep}{0pt}
\item clinical frequency (e.g., qd (once a day), bid (twice a day))
\item duration (e.g., `over night', `weekend', `months')
\item festival (e.g., `Yom Kippur', `Nowruz', `Christmas')
\item season (e.g., `summer', `spring', `autumn', etc.)
\item weekday (e.g., `Monday', `Tuesday', `Wednesday', etc.)
\item month (e.g., `January', `February', `March', etc.)
\item literal time (e.g., `morning', `afternoon', `evening')
\item temporal modifier (e.g., `on', `after', `before')
\item ordinal number (e.g., `first', `second', `third', etc.)
\item literal number (e.g., `one', `two', `three', etc.)
\end{itemize}

\subsection*{Temporal expression recognition}
The TER component consists of combined rule- and ML-based methods.\\

\textbf{The rule-based component} consist of a total of 65 rules containing patterns derived from an initial collocation extraction (i.e., bi- and tri-grams) and pattern analysis of TEs in the training data. For example, the TE patterns `MM/DD/YYYY', `MM/DD/YY', `YYYY/DD/MM' and `MM/DD' accounted for roughly 35\% of temporal expressions in the training data (i2b2-TRC).

The rule set developed combines a few types of feature: \begin{inparaenum}[(a)] \item semantic: temporal categories derived from the set of specific temporal knowledge resources during the pre-processing (see previous sub-section), \item lexical: such as common recurring expressions (e.g., `postoperative day one', `hospital day five', `today'), and \item pattern features e.g., `MM/DD/YYYY', `MM/DD/YY'.\end{inparaenum}\\

\textbf{The ML-based component} was developed using a set of features selected based on an initial literature review, and further refinement using a combination of manual forward and backward feature selection approach. A total of 19 most discriminate features were selected from an initial set of 120 features. These features can be organised into three sets:

\begin{itemize}
 \setlength{\itemsep}{1pt}
 \setlength{\parskip}{0pt}
 \setlength{\parsep}{0pt}
\item[\textbf{Lexical}] 
\item $fg_1$: the token string or alphanumeric character sequence
\item $fg_2$: semantic temporal categories derived from the `NLP pre-processing'
\item[\textbf{Orthographic}]
\item $fg_3$: token kind given by the literal representation: \textit{word}, \textit{number}, \textit{symbol}, or \textit{punctiuation}
\item $fg_4$: token-case given by the literal representation: \textit{lower-case}, \textit{upper-case}, \textit{upper-initial}, \textit{mixed-caps}, \textit{all-caps}
\item[\textbf{Combined}]
\item $fg_5$: concatenation of the features: $fg_1$, $fg_2$ and $fg_4$
\end{itemize} 

In addition to these features, the feature space consists of contextual features. Specifically, we found that the optimal feature window size of 5 or [-2,2] was ideal for $fg_1$, $fg_3$ and $fg_4$, and a window size of 3 or [0,2] for $fg_2$ (Table \ref{table:TERtemplate} gives the complete feature space used).
 
\begin{table}[h] \caption{Feature template: clinical TER} \label{table:TERtemplate}
\begin{small}
\textit{CRF feature template used for the TER. }
\end{small}
\begin{center}
\begin{tabular}{ll} \hline
\textbf{$fg_1$:Token} & \textbf{$fg_2$:Dictionary} \\
U00:\%x[-2,1] &  \\
U01:\%x[-1,1] &  \\
U02:\%x[0,1]  & U05:\%x[0,2] \\
U03:\%x[1,1]  & U06:\%x[1,2] \\
U04:\%x[2,1]  & U07:\%x[2,2] \\
& \\
\textbf{$fg_3$:TokenKind} & \textbf{$fg_4$:TokenCase} \\ 
U08:\%x[-2,5] & U13:\%x[-2,6] \\
U09:\%x[-1,5] & U14:\%x[-1,6] \\
U10:\%x[0,5]  & U15:\%x[0,6] \\
U11:\%x[1,5]  & U16:\%x[1,6] \\
U12:\%x[2,5]  & U17:\%x[2,6] \\ 
& \\
\textbf{$fg_5$:Combined} & \\
\multicolumn{2}{l}{U18:\%x[0,1]/\%x[0,2]/\%x[0,4]} \\ \hline
\end{tabular}
\end{center}
\end{table}

The ML-based module uses a a state-of-the-art sequence labelling algorithm (CRF) trained with the IO token representation schema with the following (default) CRF parameters: $C=1.00$, $ETA: 0.0001$ and L2-regularisation algorithm.

\subsubsection*{Results integration}
The output of the ML- and rule-based methods are combined at the mention level: union of the respective overlapping and non-overlapping outputs. 

\subsubsection*{Post-processing}
A rule-based post-processing component was developed in order to correct obvious and systematic errors from the hybrid TER method. This component removes common false-positives predictions identified during the development of the TER component. Common examples include single character predictions and non-related but similar numerical expressions e.g., pulmonary artery pressure measures (e.g., `42/21') and other (partial) numerical expressions such as telephone, fax and ward numbers.

\subsubsection*{TE normalisation}
The ClinicalNorMA \citep{Kovacevicetal13} is adopted as the TE normalisation component. The normaliser is based on the general domain normalisation component TRIOS \citep{Uzzaman-Allen10}. Further, ClinicalNorMA is rule-based and adheres to the TIMEX3 schema, specifically, the extended schema described in \citep{Sunetal13}. 

\section{Temporal relation identification and classification}
\label{sec:tlink_extraction}
The aim of temporal relation extraction is the chronological ordering of events. The identification of temporal links between entity pairs such as EVENTs (e.g., \textit{Problem}, \textit{Treatment}, \textit{Test}), TEs, and EVENTs and TEs, as well as the subsequent classification of these links into predefined categories (e.g., \textit{After}, \textit{Before}, \textit{Overlap}) is known as TLINK extraction.

The TLINK method developed and described herein is rule-based. The developed approach is motivated by a gap in current literature of pure knowledge driven methods for clinical TLINKs extraction (see Section \ref{sec:TIE}).

The developed method has two main components. The first component takes as input extracted clinical concepts (\textit{Problem}, \textit{Treatment}, and \textit{Test}) and TEs (\textit{Date}, \textit{Time}, \textit{Duration} and \textit{Frequency}), and generates TLINK candidate pairs (the \textit{identification} step) and subsequently \textit{classifies} the identified links into three different categories: \textit{After}, \textit{Before}, or \textit{Overlap}. A final component derives the transitive closure (refer to Appendix \ref{sec:appendic_trans_closure}) of relations extracted in order to generate implied relations that have been missed by the preceding TLINK method.

\begin{figure*}[h] 
\begin{center}
\includegraphics*[scale=0.7]{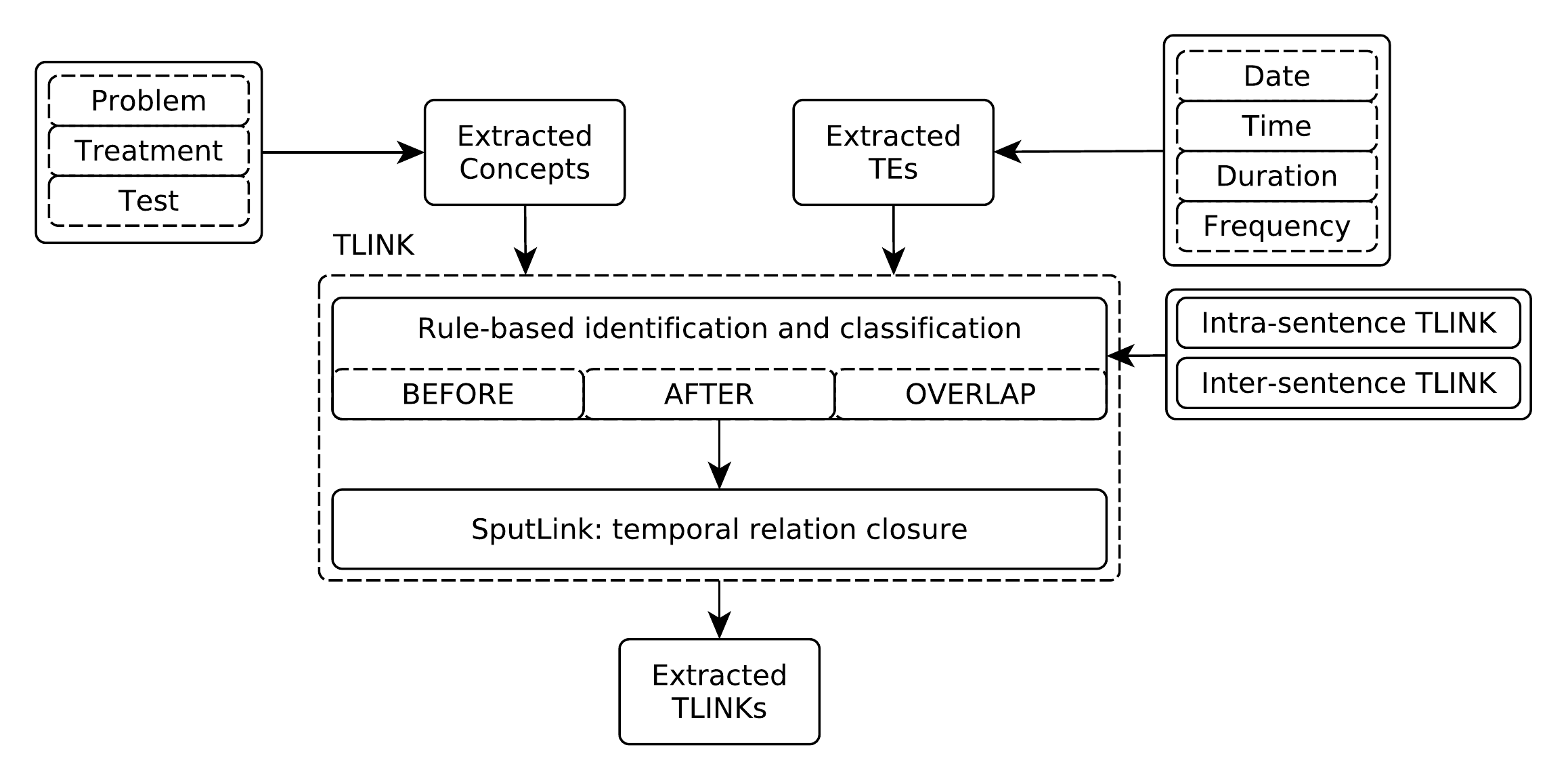} \caption{TLINK extraction architecture} 
\label{figure:TLINKarchitecture}
\end{center}
\end{figure*}

Figure \ref{figure:TLINKarchitecture} shows an abstract representation of the methodology. The remaining part of this section describes our methods in detail. 

\subsubsection*{TLINK identification and classification}
A notable difference between previous work and our approach is that we use \begin{inparaenum}[(i)] \item a pure rule-based method for TLINK extraction, and \item combine the TLINK candidate generation (identification) and classification into a single simultaneous component.\end{inparaenum}

The rule based TLINK component is partitioned into two sub-components: 
\begin{enumerate}[(1)] 
\item intra-sentence: TLINKs within a sentence span; 
\item inter-sentence: TLINKs across sentences.
\end{enumerate}

\subsubsection*{Intra-sentence TLINKs}
In order to analyse intra-sentence TLINKs, we first performed an initial semi-automatic analysis in the development dataset. For each sentence containing a TLINK, the TLINK pairs were abstracted to their respective EVENT or TIMEX3 types. Additionally, any context to the right and left of the TLINKs were removed to easily spot patterns. Subsequently, the abstracted TLINK pairs were manually analysed for common patterns by the given TLINK category. For example, in the following sentences (a, b) the underlined EVENTs and TEs are part of six different TLINKs (or three TLINKs per sentence): 

\begin{itemize}
\item[(a)] `The patient reported \underline{vomiting}, \underline{nausea} and \underline{headaches}.'
\item[(b)] `The patient received \underline{steroids} for \underline{his swelling} in \underline{2006}.'
\end{itemize}

In the following list, all pair-wise EVENTs or TE, part of TLINK is abstracted to their respective label and any context to the left and right of the pair-wise link is removed (illustrated by being \sout{strikeout}). 

\begin{itemize}
\item[($a_1$)] `\sout{The patient reported} \underline{PROBLEM}, \underline{PROBLEM} \sout{and headaches}.
\item[($a_2$)] `\sout{The patient reported} \sout{vomiting,} \underline{PROBLEM} and \underline{PROBLEM}.'
\item[($a_3$)] `\sout{The patient reported} \underline{PROBLEM}, nausea and \underline{PROBLEM}.'
\item[($b_1$)] `\sout{The patient received} \underline{TREATMENT} for \underline{PROBLEM} \sout{in 2006}.'
\item[($b_2$)] `\sout{The patient received steroids for} \underline{PROBLEM} in \underline{DATE}.'
\item[($b_3$)] `\sout{The patient received} \underline{TREATMENT} for his swellings in \underline{DATE}.'
\end{itemize}

This approach enabled us to profile various TLINK categories and formalise extraction rules based on common abstraction patterns. For example, Table \ref{table:TLINKAbsEx} lists a number of common patterns found and their typically associated TLINK category. 

\begin{table}[h] \caption{TLINK patterns} \label{table:TLINKAbsEx}
\begin{small}
\textit{This table show common patterns semi-automatically extracted from the development/training dataset. The patterns listed in this tables make up the largest and most obvious TLINK patterns observed.}
\end{small}
\begin{center}
\begin{tabular}{lcl}\hline
\multicolumn{1}{c}{\textbf{TLINK abstraction patterns}} && \multicolumn{1}{c}{\textbf{Typical TLINK}} \\ \hline
PROBLEM and PROBLEM & $\rightarrow$ & $[$PROBLEM$]$ \textit{Overlap} $[$PROBLEM \\
PROBLEM, PROBLEM & $\rightarrow$ & $[$PROBLEM$]$ \textit{Overlap} $[$PROBLEM$]$ \\
TREATMENT on DATE & $\rightarrow$ & $[$TREATMENT$]$ \textit{Overlap} $[$DATE$]$ \\
TREATMENT in DATE & $\rightarrow$ & $[$TREATMENT$]$ \textit{Overlap} $[$DAT$]$E \\
TREATMENT for PROBLEM & $\rightarrow$ & $[$TREATMENT$]$ \textit{Before} $[$PROBLEM$]$  \\
TREATMENT of PROBLEM & $\rightarrow$ & $[$TREATMENT$]$ \textit{Before} $[$PROBLEM$]$ \\ 
TEST showed PROBLEM & $\rightarrow$ & $[$TEST$]$ \textit{Before} $[$PROBLEM$]$ \\ 
PROBLEM after TREATMENT & $\rightarrow$ & $[$PROBLEM$]$ \textit{After} $[$TREATMENT$]$\\ 
TREATMENT post TEST & $\rightarrow$ & $[$TREATMENT$]$ \textit{After} $[$TEST$]$\\ 
\hline
\end{tabular}
\end{center}
\end{table}

Profiling of TLINKs revealed the occurrence of different types of relations at the sentence level which we group into three different types: \textit{co-ordinate}, \textit{prepositional}, and \textit{other} TLINKs. Further, these three types of TLINKs directly correspond to the type of extraction rules, which take advantage of corresponding features that characterised them. Specifically:

\begin{itemize}
\item \textbf{co-ordinate TLINKs} are links that are characterised by EVENTs separated by co-ordinate conjunctions such as `and', `or', or comma (i.e., `,'). For example, in the sentence (a) above, all events are co-ordinate TLINKs. In the development dataset we observed that co-ordinate TLINKs as predominately categorised as `overlap'.

\item \textbf{prepositional TLINKs} are characterised by EVENTs/TEs that are linked by a prepositions. For example, in sentence (b), the preposition `for' between the two EVENTs indicates the presence of a TLINK (in this particular example the TLINK is $[$TREATMENT$]$ after $[$PROBLEM$]$). 

\item \textbf{other TLINKs} are links that do not fit in either of the previously characterised types. A notable number of other TLINKs are characterised by linking verbs between EVENTs. For example, in the sentence `The TEST \textit{revealed} PROBLEM', TEST is linked, by the verb `revealed', to PROBLEM (in this particular example the TLINK is: $[$TEST$]$ \textit{Before} $[$PROBLEM$]$).
\end{itemize}

Table \ref{table:TLINK-features-ch5} lists and describes the type of features used to extract intra-sentence TLINKs. 

\subsubsection*{Inter-sentence TLINKs}
TLINKs that span across sentences fall into two characterised types: SECTIME and co-referential TLINKs.
\begin{itemize}

\item \textbf{SECTIME TLINKs} represent the largest proportion of inter-sentence TLINKs (e.g., in the full i2b2-TRC corpus, 45.87\% of all TLINKs are SECTIME links \citep{Sunetal13}). These are links anchored to relevant document section. Specifically, in the i2b2-TRC dataset, all events within `History of Present Illness' or related sections are linked to the admission date, and events within the `Hospital course' section are linked to discharge date. SECTIME links are predominately categorised as \textit{Before}.\\ 

Notably, it is not uncommon that clinical narratives do not contain sectime. More commonly events are anchored to the document creation time also known as DocTimeRel (document creation time relation). 

\item \textbf{Co-referential TLINKs} are EVENT co-references. These type of TLINKs are characterised as multiple EVENT mentions that refer to the same EVENT.
\end{itemize}

The approach for these two types of inter-sentence TLINKs differed. In the i2b2-TRC datasets, for development and testing, SECTIME TLINKs were addressed in a three step approach:

\begin{enumerate}[(1)]
\item extract admission and discharge dates;
\item apply Section Boundary Detection (SBD), i.e., identify `history of present illness' and `hospital course' sections accordingly;
\item anchor each EVENT in a given document section to the appropriate section date and set each TLINK category to \textit{Before}.
\end{enumerate}

However, in the case-study data there were couple notable differences to how SECTIME TLINKs were extracted. Namely, as there only existed one section time i.e., the DRD, the SBD was omitted and each EVENT was anchored to the DRD. In addition, while each TLINK category was initially set to the default link type \textit{Before}, we observed a number of common events that occurred on the DRD: routine clinical measurements such as weight, height, blood pressure, and similar. These contained TLINK type were all amended accordingly to \textit{Overlap}. 

Co-referential TLINKs are approached by considering a novel feature: lexical-level similarity (i.e., comparing literal strings with no additional features considered) between EVENTs in a given clinical note. A combined token- and character-level string similarity metric SoftTFIDF algorithm \cite{Cohenetal03} was adopted to determine the \textit{similarity} between two candidate events. Specifically, the SoftTFIDF component take as input two strings and outputs a similarity score: a real number between [0,1]; where 1 is a perfect match and 0 the vice versa. The optimum threshold of 0.8 was determined through systematic experimentation with the i2b2-TRC development set. 

The co-referential TLINK pseudo method developed is given below: 

\begin{enumerate}[(1)]
\item using SoftTFIDF, $n^{2}-1$ comparisons are done between events in a given document (including across document sections, if any);
\item if the SoftTFIDF similarity score between any pair-wise EVENTs is greater or equal to the threshold (0.8): create a TLINK between EVs with the link category: \textit{Overlap}.
\end{enumerate}

\subsubsection*{TLINKs features}

Table \ref{table:TLINK-features-ch5} list the type of features used across both intra- and inter sentence TLINK methods. The features are used as part of formalised rules and heuristics to identify and classify TLINKs and include:

\begin{table}[h] \caption{TLINK extraction features} \label{table:TLINK-features-ch5}
\begin{small}
\textit{The features listed herein were used for both TLINK identification and classification; description of each feature type follows this table. Nota bene: EV=EVENT and ST=SECTIME.}
\end{small}
\begin{center}
\begin{tabular}{|r|c|c|c|c|c|} \hline
\multicolumn{1}{|c|}{\multirow{2}{*}{\textbf{Feature type}}} & \multicolumn{3}{c|}{\textbf{Inter-sentence}} & \multicolumn{2}{c|}{\textbf{Intra-sentence}} \\ \cline{2-6} 
					& EV-EV & EV-TE & EV-ST	& EV-EV & EV-TE \\ \hline
String similarity	& 	\cmark	&		& 		&		& 		\\ \hline
Position information& 		&		& 	\cmark	&	\cmark	& 	\cmark	\\ \hline
Distance information& 		&		&		&	\cmark	& 	\cmark	\\ \hline
Preposition 		& 		&		&		&	\cmark	& 	\cmark	\\ \hline
Conjunction			& 		&		&		&	\cmark	& 	\cmark	\\ \hline
TE-related 			&	\cmark	& 		&		&		& 	\cmark	\\ \hline
NE-related	 		&	\cmark	& 		& 		&	\cmark	& 	\cmark	\\ \hline
\end{tabular}
\end{center}
\end{table}

Description of feature types listed in Table \ref{table:TLINK-features-ch5} follows.
\begin{itemize}
\item \textbf{String similarity}: specifically, string similarity score between pair-wise EVENTs derived from SoftTFIDF were used to extract co-referential TLINKs;
\item \textbf{Position information}: the position of an EVENT within a given section (SECTIME TLINKs);
\item \textbf{Distance information}: \begin{inparaenum}[(i)] \item token distance between entity pairs, and \item number of EVENT and TE between entity pairs; \end{inparaenum} 
\item \textbf{Preposition}: between two candidate pairs e.g., `in', `on', `after', `before' and so forth;
\item \textbf{Conjunction}: lexical cues between two candidate pairs e.g., `and', `both' and so forth;
\item \textbf{TE-related}: TE type i.e., date, time, duration, and frequency;
\item \textbf{EVENT-related}: EVENT information such as type i.e., \textit{Problem}, \textit{Treatment}, \textit{Test}; including HrQoL concept categories) and negation information;
\end{itemize}

\subsubsection*{Temporal links closure}
In order to capture implied TLINKs not captured by the initial rule-based method the transitive closure may be calculated. The final TLINK component engineered calculates the full set of transitive relations or temporal closure of links extracted using the initial rule-based component. Description of transitive closure is given in Appendix \ref{sec:appendic_trans_closure}. However, the explicit results including transitive closure has not been included, except the inherit evaluation provided by the TempEval-3 metric.

\section{Experiments, Results and Discussions}
\label{sec:experiments_results_discussions}

\subsection{Data}
The NER, TERN and TLINK methods presented in this report were developed and validated using a set of publicly available research datasets. The NLP research datasets used were obtained from the clinical TM challenges organised by the i2b2\footnote{the research datasets provided by i2b2 are not entirely public, and require data user agreements to be signed; \url{https://www.i2b2.org/NLP/DataSets/}.}. Specifically, these datasets are derived from the following shared-tasks: \begin{enumerate}[(i)] 
\item The 2010 i2b2 $4^{th}$ Shared Task; referred to as \textit{i2b2-CARC} hereafter \citep{Uzuneretal11}, and 
\item The 2012 i2b2 $6^{th}$ Shared Task; referred to as \textit{i2b2-TRC} hereafter \citep{Sunetal13b}.
\end{enumerate}
Table \ref{table:NLP-datasets} provides details such the size (number of documents across training and test datasets). 

\begin{table}[h] \caption{NLP datasets} \label{table:NLP-datasets}
\begin{small}
\textit{This table shows the NLP datasets used in this report.}
\end{small}
\begin{center}
\begin{tabular}{l|l|rr} \hline
\textbf{Dataset} & \textbf{Annotation} & \textbf{Training} & \textbf{Test}\\ \hline
i2b2-TRC & EVENT\footnote{Include annotated EVENTs such as \textit{Problem}, \textit{Treatment} and \textit{Test}, \textit{Occurrence}, \textit{Evidential} and \textit{Clinical department}.},TIMEX3,TLINK & 190 & 120\\ 
i2b2-CARC & EVENT\footnote{Include annotated EVENTs such as \textit{Problem}, \textit{Treatment}, \textit{Test}.} & 170 & 256\\ \hline
\end{tabular}
\end{center}
\end{table}

These datasets were produced using multiple annotators, including domain experts. Specifically, the i2b2-TRC was produced using eight expert annotators, four of whom had medical background; the i2b2-CARC was produced using twelve annotators including six with medical background\footnote{Annotation task information regarding i2b2-CARC corpus was obtained by email from responsible researcher Brett South, Senior scientist (currently) at University of Utah, Department of Biomedical Informatics.}.

\subsubsection*{EVENT}
The dataset utilised to engineer the event extraction method was composed of the i2b2-TRC and i2b2-CARC corpora. A total of 736 discharge summaries was used across the training (616 documents) and test (120 documents; i2b2-TRC test dataset). Table \ref{table:NER-corpus-stats} shows the label distribution by event/concept category across the combined datasets used in this report. 

\begin{table}[h] \caption{EVENT label distribution} \label{table:NER-corpus-stats}
\begin{small}
\textit{In this report, the i2b2-TRC (training) and i2b2-CARC (training and test) data was combined as the training dataset, while the i2b2-TRC test dataset was used as the held-out test data for the clinical NER method described herein.}
\end{small}
\begin{center}
\begin{tabular}{lrr} \hline
\textbf{EVENT} & \textbf{Training} & \textbf{Test} \\ \hline
Problem & 24,330 & 4,309 \\
Treatment & 17,773 & 3,285 \\
Test & 16,062 & 2,173 \\ \hline
\textbf{Total} & 58,165 & 9,767 \\ \hline
\end{tabular}
\end{center}
\end{table}

Table \ref{table:event-iaa-trc} show the IAA for i2b2-TRC \cite[p.808]{Sunetal13b}\footnote{These statistics are computed for \textit{Problem}, \textit{Treatment} and \textit{Test}} and Table \ref{table:event-iaa-carc} show the IAA for i2b2-CARC dataset\footnote{These statistics are computed across six different EVENTs: \textit{Problem}, \textit{Treatment}, \textit{Test}, \textit{Occurrence}, \textit{Evidential} and \textit{Clinical department}. Only the first three EVENT categories are considered in this report.}. The IAA scores confirm that recognition of event boundaries for both i2b2-TRC and i2b2-CARC is a fairly straight forward task for manual processing; with the identification of \textit{Problem}, \textit{Treatment} and \textit{Test} event boundaries being a simpler task (see Table \ref{table:event-iaa-carc}). Likewise, classification of EVENT \textit{type} and concept negation reveal to be a relatively straight forward manual annotation task for appropriately trained experts.

\begin{table}[h]
\begin{minipage}[h]{0.5\linewidth}
\centering
\caption {i2b2-TRC: EVENT IAA} \label{table:event-iaa-trc} 
\vspace{5mm}
\begin{tabular}{lccccc} 
\hline
\textbf{EVENT}	 & \multicolumn{2}{c}{\textbf{\textbf{$Avg. P\&R$}}} & \multicolumn{2}{c}{\textbf{\textbf{$\kappa$}}} \\\hline
Span (strict) & \multicolumn{2}{c}{0.83}     & \multicolumn{2}{c}{-}  \\ 
Span (lenient)& \multicolumn{2}{c}{0.87}     & \multicolumn{2}{c}{-} \\ \cdashline{1-5}
Type   	  & \multicolumn{2}{c}{0.93}   & \multicolumn{2}{c}{0.90} \\ 
Negation   	  & \multicolumn{2}{c}{0.97}   & \multicolumn{2}{c}{0.21} \\ \hline
\end{tabular}
\end{minipage}%
\qquad
\begin{minipage}[h]{0.45\linewidth}
\centering
\caption {i2b2-CARC: EVENT IAA} \label{table:event-iaa-carc} 
\vspace{5mm} 
\begin{tabular}{lccc}
\hline
\textbf{EVENT}	 & \multicolumn{2}{c}{\textbf{\textbf{$Avg. P\&R$}}}  \\\hline
Span (strict) & \multicolumn{2}{c}{0.85}   \\ 
Span (lenient)& \multicolumn{2}{c}{0.91}   \\ \hline
& \\ \hfill 
\end{tabular}
\end{minipage}
\end{table}
\newpage
\subsubsection*{TIMEX3}
The i2b2-TRC dataset was used for the development and evaluation of the TERN component. A total of 310 discharge summaries was used across the development (190 documents) and test (120 documents) datasets. Table \ref{table:temporal-corpus-stat} and Table \ref{table:timex3-iaa} show the label distribution across the dataset by TE type and the IAA, respectively \citep[p.808]{Sunetal13b}. Notably, while the IAA shows a fairly good agreement for the recognition of TE spans (with strict boundary identification proving more challenging), normalisation of TE (i.e., \textit{value}) seems even more challenging for manual efforts.

\begin{table}[h]
\begin{minipage}[h]{0.5\linewidth}
\centering
\caption {TIMEX3 label distribution} \label{table:temporal-corpus-stat} 
\vspace{5mm}
\begin{tabular}{lcccc} \hline
\textbf{Type}& 	\multicolumn{2}{r}{\textbf{Training}} & \multicolumn{2}{r}{\textbf{Test}} \\ \hline
Date  & \multicolumn{2}{r}{1,641}     & \multicolumn{2}{r}{1,222}  \\ 
Duration     & \multicolumn{2}{r}{407}      & \multicolumn{2}{r}{341}   \\ 
Frequency    & \multicolumn{2}{r}{249}      & \multicolumn{2}{r}{197}   \\ 
Time         & \multicolumn{2}{r}{69}       & \multicolumn{2}{r}{60}    \\ \hline
\textbf{Total}        & \multicolumn{2}{r}{2,366}     & \multicolumn{2}{r}{1,820}  \\ \hline
\end{tabular}
\end{minipage}%
\qquad
\begin{minipage}[h]{0.45\linewidth}
\centering
\caption {i2b2-TRC: TIMEX3 IAA} \label{table:timex3-iaa}
\vspace{5mm} 
\begin{tabular}{lccccc}
\hline
\multicolumn{1}{c}{\textbf{TIMEX3}}	 & \multicolumn{2}{c}{\textbf{\textbf{$Avg. P\&R$}}} & \multicolumn{2}{c}{\textbf{\textbf{$\kappa$}}} \\\hline
Span (strict) & \multicolumn{2}{c}{0.73}     & \multicolumn{2}{c}{-}  \\ 
Span (lenient)& \multicolumn{2}{c}{0.89}     & \multicolumn{2}{c}{-} \\ \cdashline{1-5}
Type     	& \multicolumn{2}{c}{0.90}   & \multicolumn{2}{c}{0.37} \\ 
Value      	& \multicolumn{2}{c}{0.75}   & \multicolumn{2}{c}{-} \\
Modifier   	& \multicolumn{2}{c}{0.83}   & \multicolumn{2}{c}{0.21} \\ \hline
\end{tabular}
\end{minipage}
\end{table}

\subsubsection*{TLINK}
The temporal relation component was developed and validated using the i2b2-TRC dataset. Note that only TLINKs that include EVENTs such as \textit{Problem}, \textit{Treatment}, Test and TIMEX3 have been considered. Table \ref{table:tlink-corpus-stat} and Table \ref{table:tlink-iaa} show the label distribution and the IAAs, respectively \citep[p.808]{Sunetal13b}. Notably, and comparably (i.e., versus EVENT and TE recognition tasks), it is apparent that TLINK identification is a challenging task (i.e., 0.39 in average precision-recall) for humans. However, manual effort for TLINK classification (i.e., \textit{type}) show reasonable performance.

\begin{table}[h]
\begin{minipage}[h]{0.5\linewidth}
\centering
\caption{TLINK label distribution} \label{table:tlink-corpus-stat}
\vspace{5mm} 
\begin{tabular}{lcccc}
\hline
\textbf{Type}&  \multicolumn{2}{r}{\textbf{Training}} & \multicolumn{2}{r}{\textbf{Test}}	\\ \hline
Before      & \multicolumn{2}{r}{11,981}  & \multicolumn{2}{r}{10,488}  \\ 
Overlap     & \multicolumn{2}{r}{7,276}      & \multicolumn{2}{r}{5,694} \\ 
After       & \multicolumn{2}{r}{1,415}     & \multicolumn{2}{r}{1,275}  \\ 
 \hline
\textbf{Total}  & \multicolumn{2}{r}{20,672} & \multicolumn{2}{r}{17,457}  \\ \hline
\end{tabular}
\end{minipage}%
\qquad
\begin{minipage}[h]{0.45\linewidth}
\centering
\caption {i2b2-TRC: TLINK IAA} \label{table:tlink-iaa}
\vspace{5mm} 
\begin{tabular}{lccccc}
\hline
\multicolumn{1}{c}{\textbf{TLINK}}	 & \multicolumn{2}{c}{\textbf{\textbf{$Avg. P\&R$}}} & \multicolumn{2}{c}{\textbf{\textbf{$\kappa$}}} \\\hline
Span (strict) & \multicolumn{2}{c}{0.39}     & \multicolumn{2}{c}{-}  \\ 
Span (lenient)& \multicolumn{2}{c}{-}     & \multicolumn{2}{c}{-} \\ \cdashline{1-5}
Type   	  & \multicolumn{2}{c}{0.79}   & \multicolumn{2}{c}{0.3} \\ \hline \hfill
\end{tabular}
\end{minipage}
\end{table}
\newpage
\subsection{Event extraction}
We explored a number of sequence label models: IO, BIO and W-BIO (where, W: single token word; B: beginning; I: inside; O: outside) in addition to a set of post-processing components. For the development/validation experiments we used the training data (Table \ref{table:NER-corpus-stats}) which we split into a validation training set (500 documents) and a validation test set (116 documents). 

\begin{table}[h]\caption{EVENT extraction validation test results} \label{table:NERvalidation}
\begin{small}
\textit{The validation test set results are obtained by training the models on a set of 500 documents and testing on a validation test set of 116 (shown here). The best results, horizontally or by EVENT category, are highlighted. From all models experimented, the IO model performed worst overall concept types, with strict scores being notably lower than other models (approximately 5\% across all concept categories). Further, the difference between BIO and W-BIO were minimal: the BIO models permed slightly better for the \textit{Problem} and \textit{Treatment} categories while W-BIO performed better on identifying the \textit{Test} category.}
\end{small}
\begin{center}
\begin{tabular}{lrccc} \hline
\multirow{2}{*}{\textbf{EVENT}} & \multirow{2}{*}{\textbf{Model}} & \textbf{Precision \%} & \textbf{Recall \%} & \textbf{$F_1$-measure \%} \\ 
& & \small strict$\mid$lenient & \small strict$\mid$lenient & \small strict$\mid$lenient \\ \hline

\multirow{3}{*}{Problem} 
& IO & 67.46$\mid$84.33 & 70.22$\mid$\textbf{87.78} & 68.81$\mid$86.02\\
& BIO & \textbf{73.20}$\mid$\textbf{85.95} & \textbf{74.63}$\mid$87.62 & \textbf{73.91}$\mid$\textbf{86.78}\\
& W-BIO & 72.32$\mid$85.83 & 73.54$\mid$87.28 & 72.92$\mid$86.55 \\ \hline

\multirow{3}{*}{Treatment} 
& IO & 73.63$\mid$89.36 & 70.65$\mid$\textbf{85.74} & 72.11$\mid$87.51\\
& BIO & \textbf{79.45}$\mid$90.37 & \textbf{74.70}$\mid$84.97 & \textbf{77.00}$\mid$\textbf{87.59}\\
& W-BIO & 79.41$\mid$\textbf{90.91} & 73.45$\mid$84.09 & 76.31$\mid$87.37\\ \hline

\multirow{3}{*}{Test} 
& IO & 75.00$\mid$89.20 & 72.13$\mid$\textbf{85.79} & 73.54$\mid$87.47\\
& BIO & 80.14$\mid$89.82 & 76.37$\mid$85.59 & 78.21$\mid$87.65 \\
& W-BIO & \textbf{80.72}$\mid$\textbf{90.34} & \textbf{76.50}$\mid$85.62 & \textbf{78.56}$\mid$\textbf{87.92}\\ \hline \hline

\multirow{3}{*}{\textbf{Micro score}} 
& IO & 71.31$\mid$87.13 & 70.88$\mid$\textbf{86.60} & 71.09$\mid$86.86\\
& BIO & \textbf{76.92}$\mid$88.30 & \textbf{75.13}$\mid$86.24 & \textbf{76.02}$\mid$\textbf{87.26} \\
& W-BIO & 76.67$\mid$\textbf{88.54} & 74.33$\mid$85.84 & 75.48$\mid$87.17\\ \hline

\end{tabular}
\end{center}
\end{table}

Our experiments showed that the IO models performed consistently worst compared to BIO and W-BIO, with the latter two models showing little difference (see Table \ref{table:NERvalidation}). For example, considering strict evaluation metrics, there is minimal difference between BIO and W-BIO models, while a notable difference can be observed between IO and BIO/W-BIO models (approximately 5\% micro $F_1$-measure). This suggests that BIO and W-BIO models are better suited for strict boundary identification of clinical concepts investigated compared to the IO sequence label schema. Moreover, when considering lenient evaluation scores, there is a minimal difference among all models, however, BIO and W-BIO models score consistently higher precision and $F_1$-measure while IO models score consistently higher recall.

The final evaluation or test results are presented in Table \ref{table:NERevaluation}. These are consistent with our findings during validation (Table \ref{table:NERvalidation}). As may be seen from both the validation and evaluation results, there is no notable difference between BIO and W-BIO models, except for W-BIO (\textit{Test}) which shows notably better results. 

In light of evaluation results that are comparable to IAA (Table [\ref{table:event-iaa-trc},\ref{table:event-iaa-carc}]), we have omitted detailed error analysis.  

\begin{table}[h] \caption{EVENT extraction results on the held-out test set} \label{table:NERevaluation}
\begin{small}
\textit{The results on the held-out test set showed similar trend to the validation results; the IO models have been excluded due to notably poor performance on the validation set. Further, similar to the validation results, BIO performed better on Problem and Treatment categories while W-BIO model performed best on the Test category.}
\end{small}
\begin{center}
\begin{tabular}{lrccc} \hline
\multirow{2}{*}{\textbf{EVENT}} & \multirow{2}{*}{\textbf{Model}}  & \textbf{Precision \%} & \textbf{Recall \%} & \textbf{$F_1$-measure \%} \\ 
& & \small strict$\mid$lenient & \small strict$\mid$lenient & \small strict$\mid$lenient \\ \hline

\multirow{2}{*}{Problem} 
& BIO & 81.52$\mid$90.68 & 82.62$\mid$\textbf{92.90} & 82.07$\mid$91.29\\
& W-BIO & \textbf{81.91}$\mid$\textbf{90.84} & \textbf{82.80}$\mid$91.83 & \textbf{82.35}$\mid$\textbf{91.33}\\ \hline

\multirow{2}{*}{Treatment} 
& BIO & 87.24$\mid$94.43 & \textbf{80.12}$\mid$\textbf{86.73} & 83.53$\mid$\textbf{90.42}\\
& W-BIO & \textbf{88.00}$\mid$\textbf{94.72} & \textbf{80.12}$\mid$86.24 & \textbf{83.88}$\mid$90.28\\ \hline

\multirow{2}{*}{Test} 
& BIO & 85.48$\mid$93.02 & 82.88$\mid$90.20 & 84.16$\mid$91.59\\
& W-BIO & \textbf{86.45}$\mid$\textbf{93.49} & \textbf{83.71}$\mid$\textbf{90.52} & \textbf{85.06}$\mid$\textbf{91.98}\\ \hline \hline

\multirow{2}{*}{\textbf{Micro score}} 
& BIO & 84.22$\mid$92.39 & 81.84$\mid$\textbf{89.78} & 83.01$\mid$91.07\\ 
& W-BIO & \textbf{84.85}$\mid$\textbf{92.66} & \textbf{82.10}$\mid$89.66 & \textbf{83.45}$\mid$\textbf{91.13}\\  \hline

\end{tabular}
\end{center}
\end{table}

The final models selected for the clinical NER pipeline was BIO for \textit{Problem} and \textit{Treatment}, and W-BIO for \textit{Test}. The final evaluation scores, including negation is given in Table \ref{table:NER_final}

\begin{table}[h] \caption{The clinical NER performance} \label{table:NER_final}
\begin{center}
\begin{tabular}{llcc}\hline
 & \textbf{$F_1$-micro} \% & \multirow{2}{*}{\textbf{Negation}} & \multirow{2}{*}{\textbf{Negation $\kappa$}} \\ 
 &strict$\mid$lenient&& \\\hline
\textbf{EVENT} & 83.21$\mid$91.17 & 0.93 &  0.65 \\ \hline
\end{tabular}
\end{center}
\end{table}

\subsubsection*{Impact analysis}
In order to justify the use of various features, datasets, and post-processing components, a series of impact analysis have been conducted and shown in Table \ref{table:feature_impact} (which shows the feature impact of different CRF features used), Table \ref{table:dataset_impact} (impact of datasets on the overall performance) and Table \ref{table:pp_impact} (impact of post-processing components).

Table \ref{table:feature_impact} shows the feature impact analysis by the micro score of EVENTs; lexical features have been used as the baseline. Notably, word stem have the most impact (+3\% strict and +2\% lenient $F_1$); POS and chunk features showed minimal impact on their own with the latter having a negative impact of -0.01\% lenient $F_1$. Further, while POS and chunk features adversely affect the precision, both show a positive effect on recall.

\begin{table}[h] \caption{EVENT recognition: feature impact analysis} \label{table:feature_impact}
\begin{small}
\textit{This table shows the feature impact of several CRF feature groups.}
\end{small}
\begin{center}
\begin{tabular}{l|ccc} \hline
\multirow{3}{*}{\textbf{Feature vector}} & \multicolumn{3}{c}{\textbf{EVENTs}}\\
& $P$-micro \% & $R$-micro \% & $F_1$-micro \% \\
& strict$\mid$lenient&strict$\mid$lenient &strict$\mid$lenient \\\hline
Baseline (Lexical) & 82.56$\mid$92.34 & 76.79$\mid$85.88 & 79.57$\mid$88.99 \\
Baseline $+$ Stem & 84.56$\mid$\textbf{92.96} & 81.37$\mid$89.44 & 82.93$\mid$91.17 \\
Baseline $+$ POS & 82.51$\mid$92.08 & 77.66$\mid$86.67 & 80.01$\mid$89.29\\
Baseline $+$ Chunk & 82.39$\mid$92.07 & 77.03$\mid$86.09 & 79.62$\mid$88.98 \\ \hline
\textbf{All features} & \textbf{84.43}$\mid$92.50 & \textbf{82.02}$\mid$\textbf{89.85} & \textbf{83.21}$\mid$\textbf{91.17}\\ \hline
\end{tabular}
\end{center}
\end{table}

Notably, using the i2b2-CARC corpus improved (strict and lenient) micro $F_1$-score with +17\% and +12\% (see Table \ref{table:dataset_impact}). 

\begin{table}[h] \caption{EVENT recognition: dataset impact} \label{table:dataset_impact}
\begin{small}
\textit{This table shows the impact of the different datasets used to train the CRF models.}
\end{small}
\begin{center}
\begin{tabular}{l|ccc} \hline
\multirow{3}{*}{\textbf{Dataset}} & \multicolumn{3}{c}{\textbf{EVENTs}}\\
& $P$-micro \% & $R$-micro \% & $F_1$-micro \% \\ 
& strict$\mid$lenient& strict$\mid$lenient& strict$\mid$lenient\\ \hline
i2b2-TRC & 69.03$\mid$82.97 & 63.04$\mid$75.77 & 65.90$\mid$79.20\\ 
\textbf{i2b2-TRC$+$i2b2-CARC } & \textbf{84.43}$\mid$\textbf{92.50} & \textbf{82.02}$\mid$\textbf{89.85} & \textbf{83.21}$\mid$\textbf{91.17}\\ \hline
\end{tabular}
\end{center}
\end{table}

Table \ref{table:pp_impact} shows the impact of the post-processing sub-components. For example, while the label-fixer has a adverse effect on the precision (-5\% strict and -4\% lenient), it has a positive impact on recall (+3\% strict and +5\% lenient). In addition, the label-fixer shows less than -1\% (strict) and more than +1\% (lenient) impact on the $F_1$-score. Boundary adjustment showed a positive effect on all strict metrics, and expectedly with no effect on lenient scores. The FP filter showed a slight positive impact on precision, and interestingly vice-versa on recall. 

\begin{table}[h] \caption{EVENT recognition: post-processing impact analysis} \label{table:pp_impact}
\textit{This table lists the performance impact of the various post-processing components.}
\begin{center}
\begin{tabular}{l|ccc} \hline
\multirow{3}{*}{\textbf{Component}} & \multicolumn{3}{c}{\textbf{EVENTs}}\\
& $P$-micro \% & $R$-micro \% & $F_1$-micro \% \\
& strict$\mid$lenient& strict$\mid$lenient& strict$\mid$lenient\\\hline
No post-processing & 88.09$\mid$96.06 & 77.64$\mid$84.66 & 82.54$\mid$90.00\\
Only label-fixer & 82.85$\mid$92.23 & 80.81$\mid$\textbf{89.97} & 81.82$\mid$91.09\\
Only boundary-adjustment & \textbf{89.34}$\mid$96.06 & 78.73$\mid$84.66 & \textbf{83.70}$\mid$90.00\\
Only FP filter & 89.14$\mid$\textbf{96.45} & 77.63$\mid$84.00 & 82.99$\mid$89.79\\ \hline \hline
\textbf{All post-processing} & 84.43$\mid$92.50 & \textbf{82.02}$\mid$89.85 & 83.21$\mid$\textbf{91.17}\\ \hline
\end{tabular}
\end{center}
\end{table}

\subsection{Temporal entity extraction}
We explored a number of methods in order to adopt the best approach for TER (validation results are given in Table \ref{table:validationTER}). For the ML-based method, we experimented with various sequence label schemas (i.e., IO, BIO and W-BIO). Notably, we discovered that all sequence label models explored performed relatively similar in terms of lenient scores, but W-BIO and BIO models performed notably better in terms of strict scores (e.g., 3-4\% $F_1$-measure). However, the strict rule-based method outperformed all ML models both in terms of lenient and strict scores (over 90\% lenient $F_1$-score). 

\begin{table}[h] \caption{TER validation results} \label{table:validationTER}
\begin{small}
\textit{The ML-based component was validated on the i2b2-TRC training data which was split 60/40\% for training and validation respectively. *The rule-based results shown was obtained using the whole training data.}
\end{small}
\begin{center}
\begin{tabular}{cccc} \hline
\multirow{2}{*}{\textbf{Method}} & \textbf{Precision \%} & \textbf{Recall \%} & \textbf{$F_1$-measure \%} \\
& \small strict$\mid$lenient & \small strict$\mid$lenient & \small strict$\mid$lenient \\ \hline
\multicolumn{1}{r}{IO} & 66.03$\mid$86.94 & 67.17$\mid$88.44 & 66.60$\mid$87.69 \\  
\multicolumn{1}{r}{BIO} & 71.26$\mid$87.85 & 70.95$\mid$87.47 & 71.10$\mid$87.66 \\
\multicolumn{1}{r}{W-BIO} & 71.80$\mid$87.85 & 71.49$\mid$87.47 & 71.65$\mid$87.66 \\ \cdashline{1-4}
\multicolumn{1}{r}{Rule-based*} & \textbf{78.66}$\mid$\textbf{89.64} & \textbf{80.15}$\mid$\textbf{91.34} & \textbf{79.40}$\mid$\textbf{90.48} \\ \hline
\end{tabular}
\end{center}
\end{table}

Using the official i2b2-TRC test set, we further evaluated the various ML models (using the complete training set to derive the models) and the rule-based method. In addition, we explored a number of combination the various ML models and the rule-based method (results are given in Table \ref{table:evaluationTER}).

\begin{table}[h] \caption{TER evaluation on the held-out test set} \label{table:evaluationTER}
\begin{small}
\textit{This table shows the evaluation results of various ML-, rule- and hybrid-based methods on the official i2b2-TRC test. }
\end{small}
\begin{center}
\begin{tabular}{cccc} \hline
\multirow{2}{*}{\textbf{Method}} & \textbf{Precision \%} & \textbf{Recall \%} & \textbf{$F_1$-measure \%} \\ 
& \small strict$\mid$lenient & \small strict$\mid$lenient & \small strict$\mid$lenient \\ \hline
\multicolumn{1}{r}{IO} 	& 64.42$\mid$87.10 & 66.65$\mid$90.11 & 65.51$\mid$88.58 \\
\multicolumn{1}{r}{BIO} & 67.45$\mid$86.63 & 69.56$\mid$89.34  & 68.49$\mid$87.96 \\ 
\multicolumn{1}{r}{W-BIO} & 68.22$\mid$86.47 & 68.41$\mid$86.70  & 68.31$\mid$86.58 \\ \cdashline{1-4}
\multicolumn{1}{r}{Rule-based} & \textbf{77.29}$\mid$\textbf{89.64} & 76.65$\mid$88.90 & \textbf{76.97}$\mid$89.27 \\ \cdashline{1-4}
\multicolumn{1}{r}{IO$+$Rule-based} & 72.03$\mid$86.62 & \textbf{78.41}$\mid$\textbf{94.29} & 75.09$\mid$\textbf{90.29} \\ 
\multicolumn{1}{r}{BIO$+$Rule-based} & 71.15$\mid$86.05 & 77.64$\mid$93.90 & 74.25$\mid$89.81 \\
\multicolumn{1}{r}{W-BIO$+$Rule-based} & 71.66$\mid$85.73 & 78.08$\mid$93.41 & 74.73$\mid$89.40 \\\hline
\end{tabular}
\end{center}
\end{table}

The evaluation on the held-out test set (Table \ref{table:evaluationTER}) shows a similar trend to the validation results (Table \ref{table:validationTER}) in terms of strict scores i.e., W-BIO and BIO performs notably better than IO: approximately +3\%. This indicates good generalisable methods. However, the IO model shows more notable improvement (than the validation results) in terms of $F_1$-measure over the W-BIO (+2\%) and BIO (0.62\%) models. The rule-based methods performs better than all ML models, except for the IO model's lenient recall.

We also explored a number of combinations between various ML models and the rule-based method (union of the output of each respective method) in order to discover any possible improvements. In particular, since the normalisation of TE is more important than recognition results, we are specifically interested in improved recall. The combination of the IO model and the rule-based method showed the best overall performance. A notable improvement, in terms of lenient recall, of +4.18\% and +5.39\% compared the best ML model (IO) and the rule-based method respectively, was achieved by the 
`IO+rule-based' method. Similarly, the strict recall was improved with +8.85\% and +1.76\% over the best ML model (BIO) and the rule-based method respectively. In addition, the best $F_1$-measure of 90.29\% was also achieved with the `IO+rule-based' method. As expected, the rule-based method achieved the best precision of all methods. This slightly exceeds the state-of-the-art system \citep{Sohnetal13}, which reported an $F_1$-score of 90.03\%. 

The normalisation scores reproduced using the i2b2-TRC test dataset are given in Table \ref{table:evaluationNorm}. As apparent by the `primary score' TERN task is a challenging task and an open research problem. 

\begin{table}[htbp] \caption{TE normalisation results} \label{table:evaluationNorm}
\begin{small}
\textit{This table gives the normalisation scores. The primary score is the product of the TER lenient $F_1$-measure and normalisation value accuracy and is considered the main TERN metric.}
\end{small}
\begin{center}
\begin{tabular}{cccc} \hline
\textbf{Type} & \textbf{Value} & \textbf{Modifier} & \textbf{Primary score} \\ \hline
0.8473 & 0.7044 & 0.8275 & 0.63\\ \hline
\end{tabular}
\end{center}
\end{table}

While automated recognition of TEs have shown comparable and exceeding human-level benchmark results (e.g., \citep{Sunetal13b,Uzzamanetal13}), normalisation remain a challenge|both for human and automated methods. For instance, the current state-of-the-art clinical TERN methods achieve a mere 66\% (primary score) which is just lower than the human benchmark of 66.75\% \citep{Sunetal13b}. Similarly, the state-of-the-art system \citep{Sohnetal13} achieved a 73\% accuracy for the normalised value attribute, notably lower to the human benchmark of 75\%. Regardless, these scores, either automated or human, are notably lower than common IE score of +90\% which is typically considered as good.

One of the notable challenges of TERN is the normalisation of relative expressions (e.g., `two weeks ago' `post-operative day').

\subsection{Temporal relation extraction}

\subsubsection*{Evaluation metrics}
The methods described herein have been validated using multiple evaluation methods/metrics. The main evaluation metric used in the 2012 i2b2 temporal relation challenge \citep{Sunetal13} was TempEval-3 type evaluation metrics where the `reduced graph' or redundant relations (i.e., a relation is redundant if it can be inferred from other relations) are removed. The \textit{TempEval-3 evaluation metric} used is described below:

\begin{itemize}
\item Precision: the total number of reduced system output TLINKs that can be verified in the gold standard closure divided by the total number of reduced system output TLINKs.
\item Recall: the total number reduced gold standard output TLINKs that can be verified in the system closure divided by the total number of reduced gold standard output TLINKs.
\end{itemize}

We initially developed and evaluated our method using gold standard EVENTs and TEs; the results of these experiments are shown in Table \ref{table:TLINKdevelopment} and Table \ref{table:TLINKtest}. In addition, an end-to-end evaluation where EVENTs, TEs and TLINKs are all tagged using bespoke methods (described in Sections [\ref{sec:event_extraction},\ref{sec:temporalentity_extraction},\ref{sec:tlink_extraction}] respectively) is shown in Table \ref{table:TLINKtest-e2e}.

\begin{table}[h] \caption{TLINK development set results} \label{table:TLINKdevelopment}
\begin{small}
\textit{This table shows the performance of the TLINK pipeline on the development/training dataset. We used two evaluation metrics: common precision-recall and the TempEval-3.}
\end{small}
\begin{center}
\begin{tabular}{lccc} \hline
\textbf{Evaluation setting} & \textbf{Precision \%} & \textbf{Recall \%} & \textbf{$F_1$-measure \%} \\\hline  
Customary precision-recall & 81.40 & 55.06 & 65.69 \\ 
TempEval-3 precision-recall & 80.43 & 48.05 & 62.59 \\ \hline 
\end{tabular}
\end{center}
\end{table}

As expected, fairly precision-bias results were achieved, as that was the aim during design and development. This leaves room for future work to further extend the current method in order to balance recall and to further optimise the overall score.

A direct comparison cannot be made between our results and work on the full i2b2-TRC dataset \citep{Sunetal13} due to the reason that our experiments were based on a reduced set of TLINKs. The full i2b2-TRC dataset included pairwise TLINKs of six different EVENTs, three more than used in our experiments. We did not include \textit{Occurrence}, \textit{Evidential} and \textit{Clinical department} as they were not relevant/useful for characterising patient journeys.

Nonetheless, we note the performance of the best systems evaluated on the full i2b2-TRC dataset as a point of reference. The best systems to-date, using gold annotations (for clinical EVENTs and TEs) achieved a $F_1$-measure of 69\% \citep{Tangetal13,Cherryetal13}. As previously mentioned in Chapter \ref{cha:Background}, both \cite{Tangetal13} and \cite{Cherryetal13} use complex hybrid methods with rule based components for candidate generation (i.e., TLINK identification). For classification of TLINKs, \cite{Tangetal13} uses a combination of CRF and SVM, whilst \cite{Cherryetal13} use a combination of MaxEnt and SVM for TLINK classification. In contrast, our method uses a knowledge based approach to recognise and simultaneously classify TLINKs. Our approach achieved an overall score of 62.96\% $F_1$-measure on the held-out test set (Table \ref{table:TLINKtest}). Further, considering common IE evaluation metrics, where system predictions are evaluated against manually annotated gold dataset without any further manipulation of labels, our approach achieved 65.34\% with customary and 62.96\% $F_1$-measure using TempEval-3 metrics. 

\begin{table}[h] \caption{TLINK results on the held-out test set} \label{table:TLINKtest}
\begin{small}
\textit{This table shows the results of the TLINK pipeline on the held-out test set. The results are presented using common precision-recall and the TempEval-3 evaluation metric.}
\end{small}
\begin{center}
\begin{tabular}{lcccc} \hline
\textbf{Evaluation setting} & \textbf{Precision \%} & \textbf{Recall \%} & \textbf{$F_1$-measure \%} \\\hline
Customary precision-recall & 81.51 & 54.52 & 65.34 \\
TempEval-3 precision-recall & 80.23 & 49.10 & 62.96 \\ \hline  
\end{tabular}
\end{center}
\end{table}

A comparison of results between the development (Table \ref{table:TLINKdevelopment}) and held-out test data (Table \ref{table:TLINKtest}), indicate good generalisability of the methods developed. For instance, consider the minimal variation in $F_1$-measures between the development and test set. Except a small drop in $F_1$-score when not including temporal closure (`Regular (no closure)'), the results on the test dataset were slightly better than those on the development set.

Table \ref{table:tlink_component_evaluation} shows the specific component-based evaluation of SECTIME, intra-sentence and inter-sentence TLINKs. For each component, the held-out test set has been reduced to only the relevant type of TLINKs (i.e., when evaluating SECTIME, only SECTIME links are retained). These evaluation results are obtained using the test dataset with gold annotations. 

Similar to the findings of the TLINK challenge described in \citep{Sunetal13}, we found that SECTIME TLINKs were easiest to extract (see Table \ref{table:tlink_component_evaluation}). Secondly, as expected, intra-sentence TLINKs where easier to extract than inter-sentence TLINKs (when exluding SECTIME TLINKs). Lastly, as concluded by previous work \citep{Sunetal13}, and equally applicable to our rule based approach, a better method to generate candidate pairs would be beneficial to optimise recall. 

\begin{table}[h] \caption{TLINK component based evaluation} \label{table:tlink_component_evaluation}
\begin{small}
\textit{This table shows the individual TLINK component based evaluation of the three TLINK sub-components: SECTIME, intra-sentence and inter-sentence TLINK methods. For each TLINK component evaluated the data has been reduced to only the relevant type of links.}
\end{small}
\begin{center}
\begin{tabular}{lccc} \hline
\textbf{TLINK} & \textbf{Precision \%} & \textbf{Recall \%} & \textbf{$F_1$-measure \%} \\ \hline 
SECTIME & 93.93 & 92.04 & 92.97 \\ 
Inter-sentence & 55.72 & 20.40 & 29.87 \\ 
Intra-sentence & 39.47 & 22.50 & 28.66 \\ \hline 
\end{tabular}
\end{center}
\end{table}

The component-based analysis also reinforces the conclusion that an extension of our method for recognition of candidate pairs ought to be explored. Currently, only neighbouring candidate EVENTs and co-referential inter-sentence TLINK are addressed. Extensions may include intra-sentence TLINKs that have multiple token distance in-between (e.g., first and last EVENTs in a sentence) and non co-referential inter-sentence TLINKs.

Moreover, Table \ref{table:tlink_component_evaluation} also shows the source of errors. Despite the aim of generating high precision rules, yet, it was challenging to replicate the manual effort. However, the highly inconsistent annotations (i.e., IAA: 0.39) indicate that the TLINK annotation themselves were a notable source of generated errors.  

\subsubsection*{End-to-end evaluation}
Table \ref{table:TLINKtest-e2e} shows the end-to-end evaluation: all entities are derived from bespoke methods such as clinical NER (described in Chapter \ref{cha:ConceptExtraction}), and the TERN method described in this chapter. 

\begin{table}[h] \caption{TLINK end-to-end results on the held-out test set} \label{table:TLINKtest-e2e}
\begin{small}
\textit{This table shows the results of the end-to-end system output: all annotations are derived from the bespoke clinical NER, TERN and TLINK methods described in this report thus far.}
\end{small}
\begin{center}
\begin{tabular}{lccc} \hline
\textbf{Evaluation method} & \textbf{Precision \%} & \textbf{Recall \%} & \textbf{$F_1$-measure \%} \\ \hline 
Customary precision-recall & 78.27 & 48.21 & 59.67 \\
TempEval-3 precision-recall & 76.87 & 43.05 & 55.19 \\\hline
\end{tabular}
\end{center}
\end{table}

As a point of reference, \cite{Tangetal13} achieved 62.78\% ($F_1$-measure) on the full i2b2-TRC dataset using the TempEval-3 evaluation method. Our method achieved 55.19\% using the same metric on the reduced dataset (in terms of event categories considered). Further, evaluating our method as per typical IE evaluation (i.e., against the gold set without any manipulation to the temporal graph) we achieved a $F_1$-measure of 59.67\%. 

While our methods shows good precision, an apparent limitation is the recall. We hypothesis that a better approach to candidate generation can address the latter gap.

\section{Conclusion}
\label{sec:conclusion}
This report describes a set of NLP methods to order clinical events onto a temporal space or timeline. A number of notable observation were made from the validation of these methods:

\begin{itemize}
\item EVENTs or broad clinical concept categories (i.e., \textit{Problem}, \textit{Treatment}, \textit{Test}) can be automatically extracted (using CRF) with comparable scores to human benchmark. 
\item negation of concepts can be automatically determined (using ConText negation tool with minor `tailoring') with comparable accuracy to human benchmark.
\item temporal entity identification can be automatically extracted  with comparable score to human benchmarks.
\item temporal entity normalisation is comparably challenging (even for humans). Further, determining the value (ISO-8601) was harder than type identification.  
\item TLINK extraction is overall an open research problem. DocTimeRel or SECTIME links can be extracted with good scores (93\%) while intra- and inter-sentence links are notably more challenging to extract. 
\end{itemize}

In future work, we will investigate the expansion of lexical features by incorporating lexical variant generation for EVENT extraction. The expansion of the TE normalisation component is currently being achieved. Additionally, expanding candidate generation heuristics and integrating machine learning classifiers are currently being investigated to improve the TLINK component.

\clearpage
\section{Apppendice}
\section*{Appendix A: Event extraction}

\subsection*{Event conceptualisation}
Table \ref{table:event-definition} shows the semantic definition of relevant event categories. 

\begin{table}[!h] \caption{Definition of EVENT categories} \label{table:event-definition}
\begin{small}
The definition of event categories are described according to the annotation guidelines 
\end{small}
\begin{center}
\begin{tabular}{l|ll} \hline
\textbf{EVENT} & \textbf{Semantic type} & \textbf{Semantic group} \\ \hline
\multirow{12}{*}{\textit{Problem}} & acquired abnormality & \multirow{10}{*}{Disorders} \\
& anatomical abnormality &  \\
& cell or molecular dysfunction &  \\
& congenital abnormality &  \\
& disease or syndrome &  \\
& injury or poisoning & \\
& mental or behavioural dysfunction &  \\
& neoplastic process & \\
& pathologic functions &  \\
& sign or symptom &  \\ \cdashline{2-3}
& bacterium &  \multirow{2}{*}{Living Beings}\\
& virus &  \\ \hline
\multirow{8}{*}{\textit{Treatment}} & antibiotic &  \multirow{5}{*}{Chemicals \& Drugs} \\
& biomedical or dental material & \\
& clinical drug &  \\
& pharmacologic substance &  \\
& steroid & \\ \cdashline{2-3}
& drug delivery device &  \multirow{2}{*}{Devices}\\ 
& medical device &  \\ \cdashline{2-3}
& therapeutic or preventive procedure &  Procedures \\  \hline
\multirow{2}{*}{\textit{Test}} & diagnostic procedure  & \multirow{2}{*}{Procedures} \\
& laboratory procedure & \\  \hline
\end{tabular}
\end{center}
\end{table}

\clearpage

\section*{Appendix B: Transitive closure: an example}
\label{sec:appendic_trans_closure}

\textbf{Transitive closure} of a given relations computes all implicit relations or take into account its transitivity. Further, we can define a transitive relation as: 
\begin{equation}
\forall_{a,b,c} \subseteq X : (a R b \wedge b R c) \Rightarrow a R c
\end{equation}

For example, in Table \ref{table:transitive-closure} the letters ${A, B, C}$ represent different EVENTs, and the arrows `$\rightarrow$', `$\leftarrow$', and `$\leftrightarrow$' represent the temporal relations `before', `after', and `overlap' respectively. Hence, given the TLINKs: EVENT $A$ before EVENT $B$, EVENT $B$ after EVENT $C$, and EVENT $A$ overlap EVENT $C$ are represent as followed $A \rightarrow B$, $B \leftarrow C$, and $A \leftrightarrow C$ respectively. 

\begin{table}[h] \caption{Transitive relations}
\label{table:transitive-closure}
\begin{small}
\textit{This table shows a number of example of transitive relations.}
\end{small}
\begin{center}
\begin{tabular}{c} \hline
If A $\rightarrow$ B and B $\rightarrow$ C, then A $\rightarrow$ C \\
If A $\leftarrow$ B and B $\leftarrow$ C, then A $\leftarrow$ C \\
If A $\leftrightarrow$ B and B $\leftrightarrow$ C, then A $\leftrightarrow$ C \\
If A $\rightarrow$ B and B $\leftrightarrow$ C, then A $\rightarrow$ C \\
If A $\rightarrow$ B and A $\leftrightarrow$ C, then C $\rightarrow$ B \\ \hline
\end{tabular}
\end{center}
\end{table}
\clearpage




\begin{thebibliography}{}
\expandafter\ifx\csname url\endcsname\relax
  \def\url#1{\texttt{#1}}\fi
\expandafter\ifx\csname urlprefix\endcsname\relax\def\urlprefix{URL }\fi
\expandafter\ifx\csname href\endcsname\relax
  \def\href#1#2{#2} \def\path#1{#1}\fi

\end{thebibliography}


\begin{thebibliography}{10}
\expandafter\ifx\csname url\endcsname\relax
  \def\url#1{\texttt{#1}}\fi
\expandafter\ifx\csname urlprefix\endcsname\relax\def\urlprefix{URL }\fi
\expandafter\ifx\csname href\endcsname\relax
  \def\href#1#2{#2} \def\path#1{#1}\fi

\bibitem{Uzuneretal11}
O.~Uzuner, B.~R. South, S.~Shen, S.~L. DuVall,
  \href{http://dx.doi.org/10.1136/amiajnl-2011-000203}{{2010 i2b2/VA challenge
  on concepts, assertions, and relations in clinical text}}, Journal of the
  American Medical Informatics Association 18~(5) (2011) 552--556.
\newblock \href {http://dx.doi.org/10.1136/amiajnl-2011-000203}
  {\path{doi:10.1136/amiajnl-2011-000203}}.
\newline\urlprefix\url{http://dx.doi.org/10.1136/amiajnl-2011-000203}

\bibitem{Sunetal13b}
W.~Sun, A.~Rumshisky, O.~Uzuner,
  \href{http://dx.doi.org/10.1136/amiajnl-2013-001628}{{Evaluating temporal
  relations in clinical text: 2012 i2b2 Challenge}}, Journal of the American
  Medical Informatics Association 20~(5) (2013) 806--813.
\newblock \href {http://dx.doi.org/10.1136/amiajnl-2013-001628}
  {\path{doi:10.1136/amiajnl-2013-001628}}.
\newline\urlprefix\url{http://dx.doi.org/10.1136/amiajnl-2013-001628}

\bibitem{Verhagenetal07}
M.~Verhagen, R.~Gaizauskas, F.~Schilder, M.~Hepple, G.~Katz, J.~Pustejovsky,
  \href{http://www.aclweb.org/anthology/S/S07/S07-1014}{{SemEval-2007 Task 15:
  TempEval Temporal Relation Identification}}, in: Proceedings of the Fourth
  International Workshop on Semantic Evaluations (SemEval-2007), Association
  for Computational Linguistics, Prague, Czech Republic, 2007, pp. 75--80.
\newline\urlprefix\url{http://www.aclweb.org/anthology/S/S07/S07-1014}

\bibitem{Verhagenetal10}
M.~Verhagen, R.~Saur\'{\i}, T.~Caselli, J.~Pustejovsky,
  \href{http://portal.acm.org/citation.cfm?id=1859674}{{SemEval-2010 task 13:
  TempEval-2}}, in: Proceedings of the 5th International Workshop on Semantic
  Evaluation, SemEval '10, Association for Computational Linguistics,
  Stroudsburg, PA, USA, 2010, pp. 57--62.
\newline\urlprefix\url{http://portal.acm.org/citation.cfm?id=1859674}

\bibitem{Uzzamanetal13}
N.~UzZaman, H.~Llorens, L.~Derczynski, M.~Verhagen, J.~Allen, J.~Pustejovsky,
  {SemEval-2013 Task 1: TEMPEVAL-3: Evaluating Time Expressions, Events, and
  Temporal Relations} (2013).

\bibitem{Bethard2015}
S.~Bethard, L.~Derczynski, J.~Pustejovsky, M.~Verhagen, {SemEval-2015 Task 6:
  Clinical TempEval}, in: 9th International Workshop on Semantic Evaluation
  (SemEval 2015), Association for Computational Linguistics, 2015.

\bibitem{Kovacevicetal13}
A.~Kovacevic, A.~Dehghan, M.~Filannino, J.~A. Keane, G.~Nenadic,
  \href{http://www.ncbi.nlm.nih.gov/pubmed/23605114}{{Combining rules and
  machine learning for extraction of temporal expressions and events from
  clinical narratives.}}, Journal of the American Medical Informatics
  Association : JAMIA 20~(5) (2013) 859--66.
\newline\urlprefix\url{http://www.ncbi.nlm.nih.gov/pubmed/23605114}

\bibitem{Harkemaetal09}
H.~Harkema, J.~N. Dowling, T.~Thornblade, W.~W. Chapman,
  \href{http://www.sciencedirect.com/science/article/pii/S1532046409000744}{{C%
onText: An algorithm for determining negation, experiencer, and temporal status
  from clinical reports}}, Journal of Biomedical Informatics 42~(5) (2009)
  839--851.
\newblock \href {http://dx.doi.org/http://dx.doi.org/10.1016/j.jbi.2009.05.002}
  {\path{doi:http://dx.doi.org/10.1016/j.jbi.2009.05.002}}.
\newline\urlprefix\url{http://www.sciencedirect.com/science/article/pii/S15320%
46409000744}

\bibitem{Uzzaman-Allen10}
N.~UzZaman, J.~Allen, \href{http://www.aclweb.org/anthology/S10-1062}{{TRIPS
  and TRIOS System for TempEval-2: Extracting Temporal Information from Text}},
  in: Proceedings of the 5th International Workshop on Semantic Evaluation,
  Association for Computational Linguistics, Uppsala, Sweden, 2010, pp.
  276--283.
\newline\urlprefix\url{http://www.aclweb.org/anthology/S10-1062}

\bibitem{Sunetal13}
W.~Sun, A.~Rumshisky, O.~Uzuner,
  \href{http://www.ncbi.nlm.nih.gov/pubmed/23564629}{{Evaluating temporal
  relations in clinical text: 2012 i2b2 Challenge.}}, Journal of the American
  Medical Informatics Association : JAMIA 20~(5) (2013) 806--13.
\newblock \href {http://dx.doi.org/10.1136/amiajnl-2013-001628}
  {\path{doi:10.1136/amiajnl-2013-001628}}.
\newline\urlprefix\url{http://www.ncbi.nlm.nih.gov/pubmed/23564629}

\bibitem{Cohenetal03}
W.~W. Cohen, P.~Ravikumar, S.~E. Fienberg,
  \href{http://citeseerx.ist.psu.edu/viewdoc/summary?doi=10.1.1.14.3605}{{A
  comparison of string distance metrics for name-matching tasks}} (2003).
\newline\urlprefix\url{http://citeseerx.ist.psu.edu/viewdoc/summary?doi=10.1.1%
.14.3605}

\bibitem{Sohnetal13}
S.~Sohn, K.~B. Wagholikar, D.~Li, S.~R. Jonnalagadda, C.~Tao, R.~{Komandur
  Elayavilli}, H.~Liu,
  \href{http://dx.doi.org/10.1136/amiajnl-2013-001622}{{Comprehensive temporal
  information detection from clinical text: medical events, time, and TLINK
  identification}}, Journal of the American Medical Informatics Association
  20~(5) (2013) 836--842.
\newblock \href {http://dx.doi.org/10.1136/amiajnl-2013-001622}
  {\path{doi:10.1136/amiajnl-2013-001622}}.
\newline\urlprefix\url{http://dx.doi.org/10.1136/amiajnl-2013-001622}

\bibitem{Tangetal13}
B.~Tang, Y.~Wu, M.~Jiang, Y.~Chen, J.~C. Denny, H.~Xu,
  \href{http://dx.doi.org/10.1136/amiajnl-2013-001635
  http://www.ncbi.nlm.nih.gov/pubmed/23571849}{{A hybrid system for temporal
  information extraction from clinical text.}}, Journal of the American Medical
  Informatics Association : JAMIA 20~(5) (2013) 828--35.
\newblock \href {http://dx.doi.org/10.1136/amiajnl-2013-001635}
  {\path{doi:10.1136/amiajnl-2013-001635}}.
\newline\urlprefix\url{http://dx.doi.org/10.1136/amiajnl-2013-001635
  http://www.ncbi.nlm.nih.gov/pubmed/23571849}

\bibitem{Cherryetal13}
C.~Cherry, X.~Zhu, J.~Martin, B.~de~Bruijn,
  \href{http://dx.doi.org/10.1136/amiajnl-2013-001624}{{\`{A} la Recherche du
  Temps Perdu: extracting temporal relations from medical text in the 2012 i2b2
  NLP challenge}}, Journal of the American Medical Informatics Association
  20~(5) (2013) 843--848.
\newblock \href {http://dx.doi.org/10.1136/amiajnl-2013-001624}
  {\path{doi:10.1136/amiajnl-2013-001624}}.
\newline\urlprefix\url{http://dx.doi.org/10.1136/amiajnl-2013-001624}

\end{thebibliography}
\section*{References}
\bibliographystyle{elsarticle-num} 

\end{document}